\documentclass{article}
\usepackage{spconf,amsmath,graphicx,hyperref}
\usepackage{xcolor}
\usepackage{soul}
\usepackage{booktabs}
\usepackage{makecell}
\usepackage{amssymb}
\usepackage{array}
\usepackage[normalem]{ulem} 
\usepackage{caption}
\usepackage{balance}   
\usepackage{placeins}

\title{RhythmBERT: A Self-Supervised Language Model Based on Latent Representations of ECG Waveforms for Heart Disease Detection}
\name{Xin Wang\textsuperscript{1}, 
      Burcu Ozek\textsuperscript{2}, 
      Aruna Mohan\textsuperscript{2}, 
      Amirhossein Ravari\textsuperscript{3}, 
      Or Zilbershot\textsuperscript{2}\sthanks{Corresponding authors.},
      Fatemeh Afghah\textsuperscript{4}\footnotemark[\value{footnote}]}
\address{\textsuperscript{1}Virginia Tech, VA, USA, 
         \textsuperscript{2}Walkky LLC, Boston, MA, USA,\\
         \textsuperscript{3}Northeastern University, Boston, MA, USA,
         \textsuperscript{4}Clemson University, Clemson, SC, USA\\
         \{xin, burcu, aruna, amir, or, fatemeh\}@walkky.com}

\begin{document}

%
\maketitle
\begin{abstract}
Electrocardiogram (ECG) analysis is crucial for diagnosing heart disease, but most self-supervised learning methods treat ECG as a generic time series, overlooking physiologic semantics and rhythm-level structure. Existing contrastive methods utilize augmentations that distort morphology, whereas generative approaches employ fixed-window segmentation, which misaligns cardiac cycles. To address these limitations, we propose RhythmBERT, a generative ECG language model that considers ECG as a language paradigm by encoding P, QRS, and T segments into symbolic tokens via autoencoder-based latent representations. These discrete tokens capture rhythm semantics, while complementary continuous embeddings retain fine-grained morphology, enabling a unified view of waveform structure and rhythm. RhythmBERT is pretrained on approximately 800,000 unlabeled ECG recordings with a masked prediction objective, allowing it to learn contextual representations in a label-efficient manner. Evaluations show that despite using only a single lead, RhythmBERT achieves comparable or superior performance to strong 12-lead baselines. This generalization extends from prevalent conditions such as atrial fibrillation to clinically challenging cases such as subtle ST-T abnormalities and myocardial infarction. Our results suggest that considering ECG as structured language offers a scalable and physiologically aligned pathway for advancing cardiac analysis.
\end{abstract}
%
\section{Introduction}
\label{sec:intro}

\begin{figure*}[t]
  \centering
  \includegraphics[width=0.95\textwidth]{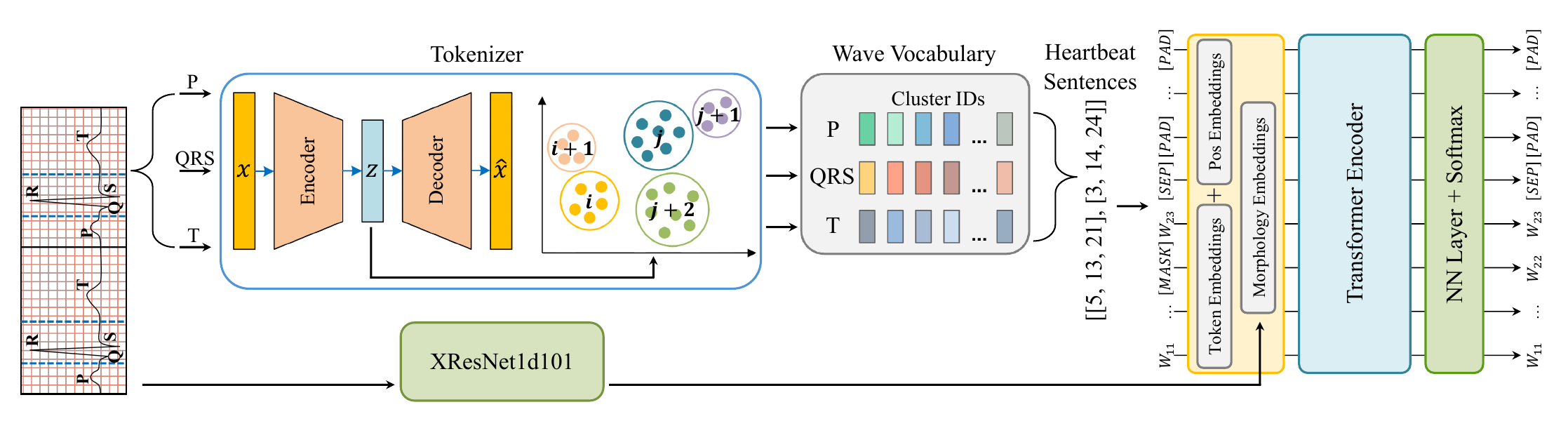}
  \vspace{-10pt}
  \caption{An overview of our proposed RhythmBERT.}
  \label{fig:res}
  \vspace{-10pt}
\end{figure*}

Cardiovascular disease remains the leading cause of mortality worldwide, making scalable and reliable electrocardiogram (ECG) interpretation a pressing challenge. ECG self-supervised learning (SSL) has emerged as a compelling way to reduce dependence on scarce expert labels by pretraining on large unlabeled corpora \cite{liu2021deep}, enabling models to acquire transferable structure directly from data.

Modern SSLs for ECG representation learning primarily follow two paradigms: contrastive and generative. Contrastive methods \cite{liu2023self, kiyasseh2021clocs} learn by maximizing agreement between augmented views of the same signal. However, their performance is critically dependent on augmentation design, as improper transforms can create ``bad positive pairs'' that obscure key features like wave morphology or inter-beat intervals \cite{lan2023towards}. By contrast, generative models like masked autoencoder learn by reconstructing masked segments of the native ECG. This approach avoids aggressive, semantics-altering augmentations and thus better preserves the physiologic integrity of the waveform's morphology and rhythm \cite{sawano2024applying, zhou2023masked}. A pioneer model, ECGBERT \cite{choi2023ecgbert}, performs masked reconstruction; however, its vocabulary is formed by using raw, high-dimensional ECG segments, which introduces a computational bottleneck and severely limits scalability.

Despite these advances, many SSLs still treat ECGs as generic time series, segmenting signals into fixed-size windows and optimizing reconstruction or contrastive objectives \cite{song2024foundation}. This misalignment with cardiac-cycle boundaries obscures both form-level (wave morphology) and rhythm-level (sequence) semantics. For example, ST-MEM \cite{na2024guiding} patchifies multi-lead ECGs and reconstructs masked patches with lead-aware cues, but its fundamental unit is a fixed-size patch rather than a physiological event. More recent work, HeartLang \cite{jin2025reading}, advances the ECG-as-language paradigm by treating heartbeats as ``words'' and rhythms as ``sentences''. However, this model is based on QRS waves only. While capturing rhythms, it does not model P- and T-wave morphology explicitly, which imposes an inductive bias that can attenuate atrial/repolarization cues and limit clinical generalization.

In this work, we propose RhythmBERT, a generative self-supervised ECG language model that captures biosemantic representations of ECG. It forms tokens via autoencoder (AE)-based waveform tokenizers. Each detected P/QRS/T segment is encoded into a latent vector; the embeddings are clustered, and the cluster ID serves as the discrete token. Thus, an ECG becomes a sequence of waveform tokens, and each cardiac cycle a grammatical “sentence”. To mitigate fidelity loss from discretization, a 1D-ResNet branch processes raw segments, and its continuous morphology embeddings are fused with the discrete tokens to preserve fine morphology. We pretrain with masked-token prediction to learn contextualized representations that respect both waveform morphology and rhythm. Across multiple downstream classification tasks, RhythmBERT demonstrates competitive and robust performance. An overview of RhythmBERT is shown in Fig.~\ref{fig:res}.

Our contributions are summarized as follows. (1) We introduce RhythmBERT, a novel, scalable self-supervised ECG language model, which is pretrained on 800,000 ECG recordings that yield strong generalization across diverse downstream tasks.  (2) We design AE-based tokenizers that encode each P/QRS/T wave into a low-dimensional latent vector via AE, which facilitates scalable, stable clustering into discrete tokens and better captures waveform morphology. (3) We fuse 1D-ResNet embeddings of raw segments with discrete tokens to reduce discretization loss and retain high-fidelity morphology. (4) We evaluate RhythmBERT on multiple downstream tasks. Notably, despite being trained exclusively on a single lead, RhythmBERT achieves performance comparable, and in some cases superior, to state-of-the-art 12-lead models. This includes strong performance on both common conditions like atrial fibrillation and diagnostically challenging cases such as subtle ST-T abnormalities, underscoring the model's robust transferability across clinically diverse scenarios. 
\vspace{-5pt}
\section{Methodology}
\label{sec:methodology}

This section details the architecture of RhythmBERT, which operates exclusively on Lead II of the ECG signal. We adopt this single-lead approach because Lead II offers the clearest view of the heart's conduction pathway and is the clinical standard for rhythm analysis. This focus is well-supported by prior research, which has demonstrated that single-lead models can achieve cardiologist-level performance \cite{tateno2001automatic, hannun2019cardiologist, silva2025systematicreviewecgarrhythmia} and have applicability to wearable devices for remote monitoring \cite{WalkkyEMBC2024}. 
\vspace{-3pt}
\subsection{ECG Waveform Tokenizer}
\label{ssec:ELP}
To treat ECG signals as sentences \cite{mousavi2021ecg}, we first preprocess the raw signals. The pipeline involves a 0.5 Hz high-pass Butterworth filter and a 50 Hz notch filter for cleaning. Following R-peak detection, a discrete wavelet transform (DWT)–based delineation algorithm \cite{martinez2004wavelet} robustly segments the P, QRS, and T waves by identifying their onsets, peaks, and offsets. These segmented waves then serve as inputs to the waveform tokenizer. The waveform tokenizer consists of three independent modules, each tasked with learning a discrete, symbolic representation for a specific waveform type (P, QRS, or T). This is achieved in a two-stage process. First, a one-dimensional convolutional autoencoder (1D-CAE) maps a raw waveform segment into a fixed-size continuous latent vector. Second, the latent vectors for each wave type are clustered to form a wave vocabulary. After that, the discrete tokens are concatenated in the order of P-QRS-T to form the heartbeat sentences. 

\noindent\textbf{\underline{Autoencoder.}} 
Each autoencoder maps a variable-length waveform segment to a fixed-length latent and reconstructs it: \vspace{-3pt}
\begin{equation}
z_n^\tau = f_\theta^\tau(x_n^\tau), \qquad \hat{x}_n^\tau = g_\phi^\tau(z_n^\tau), \qquad z_n^\tau \in \mathbb{R}^{d_\tau}.
\end{equation}
\noindent The encoder $f_\theta^\tau$ is parameterized by $\theta$, the decoder $g_\phi^\tau$ by $\phi$, and $d_\tau$ is the latent dimension. 
Training minimizes the Huber reconstruction loss (Smooth L1, $\beta=1.0$): \vspace{-3pt}
\begin{equation}
\mathcal{L}_\mathrm{AE}^{\tau}
=\frac{1}{N_\tau}\sum_{i=1}^{N_\tau}\frac{1}{T_{i,\tau}}
\sum_{t=1}^{T_{i,\tau}} 
\ell_\beta\!\left(x_i^\tau[t] - \hat{x}_i^\tau[t]\right),
\end{equation}
where 
$
\ell_\beta(r) = \tfrac{1}{2} r^2 \,\mathbf{1}_{|r|\le \beta} \;+\; \beta \big(|r| - \tfrac{1}{2}\beta\big)\,\mathbf{1}_{|r|>\beta}.
$
\noindent Here, $x_i^\tau \in \mathbb{R}^{T_{i,\tau}}$ is the $i$-th segment of type $\tau\!\in\!\{\mathrm{P},\mathrm{QRS},\mathrm{T}\}$ with length $T_{i,\tau}$, 
$z_i^\tau \in \mathbb{R}^{d_\tau}$ its latent representation, 
$\hat{x}_i^\tau$ its reconstruction, 
and $\mathcal{D}_\tau=\{x_i^\tau\}_{i=1}^{N_\tau}$ the training set for wave type $\tau$.
Each autoencoder consists of 4 convolutional blocks, each containing a 1D convolution layer, followed by batch normalization \cite{ioffe2015batch} and a GELU activation function \cite{hendrycks2016gaussian}. The output channels for the four sequential blocks are {32, 64, 128, 256}. This stride architecture reduces the temporal resolution by a factor of $2^4$. Following the final convolutional block, the feature map is processed by parallel 1D adaptive average and max pooling layers. The outputs of these layers are concatenated, flattened, and projected by a linear layer into the final latent space. The dimensionality of the latent vectors was empirically set to 12 for relatively simple P and T waves, and 24 for QRS complexes. The decoder symmetrically mirrors the encoder's architecture, employing 1D transposed convolutional layers to reconstruct the original input waveform.

\noindent\textbf{\underline{Clustering.}} Following dimensionality reduction by the autoencoder, we quantize the latent vectors into a discrete wave vocabulary using k-means clustering. 
We evaluated several clustering methods, including k-means, hierarchical, density-based, and matrix profile distance-based clustering, and selected k-means clustering due to its robust performance and computational efficiency. To determine the optimal number of clusters ($k$) for each wave type, we employ the elbow method. This involves calculating the Within-Cluster Sum of Squares (WCSS) for a range of potential $k$ values, with the optimal number identified using the KneeLocator \cite{satopaa2011finding} algorithm at the point of maximum curvature on the WCSS plot. The optimal $k$ of the P, QRS, and T waves is 13, 11, and 10, respectively. The final partitioning uses k-means++ \cite{arthur2006k} initialization, and cluster quality is validated with silhouette scores \cite{rousseeuw1987silhouettes}. 

This procedure partitions the latent vectors, and each wave segment is subsequently represented by its assigned cluster ID. Finally, these numerical IDs are concatenated in physiological order (P-QRS-T) to formulate ``heartbeat sentences", creating structured data for the BERT model.
\vspace{-4pt}
\subsection{BERT Backbone Network}
\label{ssec:Backbone}
\vspace{-2pt}
\setlength{\abovecaptionskip}{-5pt}
\begin{table}[b]
\caption{Labels used in the three downstream datasets.}\vspace{2pt}
\label{tab:labels}
\centering
\small
\setlength{\tabcolsep}{5pt}
\renewcommand{\arraystretch}{1.15}
\begin{tabular}{@{} >{\raggedright\arraybackslash}m{0.40\linewidth} >{\raggedright\arraybackslash}m{0.58\linewidth} @{}}
\toprule
\textbf{Dataset} & \textbf{Labels} \\
\midrule
\textbf{PTB-XL} & NORM, MI, STTC, HYP, CD \\
\textbf{CPSC2018} & \makecell[l]{NORMAL, AF, 1AVB, LBBB, RBBB,\\ PAC, PVC, STD, STE} \\
\textbf{Chapman--Shaoxing} & \makecell[l]{SB, SR, AFIB, ST, AF, SA, SVT,\\AT, AVNRT, AVRT, SAAWR} \\
\bottomrule
\end{tabular}
\end{table} 
\setlength{\belowcaptionskip}{-10pt}
The core of our framework is a transformer encoder designed to learn deep contextual representations from the symbolic heartbeat sentences. Each heartbeat sentence, a sequence of three waveform tokens (P, QRS, T), is appended with a [SEP] token to mark the end of the sequence. The sentences are then structured for batch processing using [PAD] tokens to ensure uniform length.

\noindent\textbf{\underline{Input Embeddings.}} (1) Token embeddings: Each discrete token ID from the heartbeat sentences is mapped to a dense vector of dimension $d_{\text{model}}$ using a trainable embedding matrix. (2) Positional Embeddings: To incorporate sequence order, learned absolute positional embeddings are added to each token embedding. (3) Morphological Embeddings: To re-inject fine-grained morphological detail from the raw signal, we extract a continuous feature vector for each waveform. These features are generated by a pre-trained XResNet1d-101, an architecture that has shown competitive performance for ECG analysis \cite{strodthoff2020deep}. 
The resulting 512-dimensional feature vector is then projected to the model's hidden dimension $d_{\text{model}}$ of 196 via a dedicated linear layer and summed with the other embeddings.

\noindent\textbf{\underline{Transformer Encoder.}} The encoder is a stack of 8 transformer layers. Each layer consists of two sub-layers: a multi-head self-attention mechanism and a position-wise feed-forward network (FFN). We use 12 attention heads and a model dimension $d_{\text{model}}$. The FFN has an inner-layer dimensionality of 4 $\times$ $d_{\text{model}}$ with a ReLU activation function. Residual connections and layer normalization are applied after each sub-layer, and dropout is used for regularization. 
\vspace{-2pt}
\section{Experiments and Analysis} 
\label{sec:experiments_analysis}
\subsection{Pretraining Configuration}
\vspace{-2pt}
\noindent\textbf{\underline{MIMIC-IV-ECG.}} A large, publicly available dataset \cite{gow2023mimic} comprises 800,035 12-lead ECGs from 161,352 patients. Each ECG recording was sampled at 500 Hz and lasted for 10 seconds. This corpus was used exclusively for the unsupervised pre-training of our model.

\noindent\textbf{\underline{Implementation.}} Our pre-training process consists of two sequential stages: waveform representation learning via autoencoders and masked token reconstruction via a transformer encoder. In the first stage, we train a separate convolutional autoencoder for each waveform type (P, QRS, T) to learn a compact latent representation. Each autoencoder is trained for a maximum of 30 epochs with a batch size of 64 per GPU. We use the AdamW optimizer with a learning rate of $1 \times 10^{-4}$ and a weight decay of $1 \times 10^{-4}$. The optimization objective is to minimize the Huber loss (also known as Smooth L1 Loss with $\beta=1.0$) for the reconstruction of the input and output waveforms. In the second stage, we pre-train the transformer encoder using a Masked Language Modeling (MLM) objective \cite{devlin2019bert}. We randomly mask 20\% of the input tokens, and the model is trained to predict the original tokens from the context. This stage is trained for a maximum of 100 epochs and a batch size of 64 per GPU. Optimization is performed using the AdamW optimizer with a weight decay of $3 \times 10^{-4}$ and a cosine annealing learning rate schedule. The schedule includes a 10\% warmup period and reaches a peak learning rate of $1 \times 10^{-4}$. The training objective is to minimize the Cross-Entropy loss. Early stopping is employed in both stages.  
\begin{table*}[t]
\caption{AUROC results of RhythmBERT and other eSSL methods. Unlike ST-MEM~\cite{na2024guiding} and HeartLang~\cite{jin2025reading}, which use 12-lead ECGs, our method relies solely on Lead II. Best results are shown in \textbf{bold}; second-best are highlighted in \colorbox{gray!25}{gray}.}
\vspace{-1pt}
\label{tab:results-auroc}
\centering
\small
\setlength{\tabcolsep}{7pt}
\begin{tabular}{lccc ccc ccc}
\toprule
\textbf{Model} &
\multicolumn{3}{c}{\textbf{PTB-XL}} &
\multicolumn{3}{c}{\textbf{CPSC2018}} &
\multicolumn{3}{c}{\textbf{Chapman--Shaoxing}} \\
\cmidrule(lr){2-4}\cmidrule(lr){5-7}\cmidrule(lr){8-10}
& \textbf{1\%} & \textbf{10\%} & \textbf{100\%}
& \textbf{1\%} & \textbf{10\%} & \textbf{100\%}
& \textbf{1\%} & \textbf{10\%} & \textbf{100\%} \\
\midrule
ST-MEM (~\cite{na2024guiding}, 12 leads)   & 61.12 & 66.87 & 71.36 & 56.69 & 63.32 & 70.39 & 59.77 & 66.87 & 71.36 \\
HeartLang (~\cite{jin2025reading}, 12 leads) & \textbf{78.94} & \textbf{85.59} & \textbf{87.52} & \colorbox{gray!25}{60.44} & \colorbox{gray!25}{66.26} & \colorbox{gray!25}{77.87} & \colorbox{gray!25}{57.94} & \colorbox{gray!25}{68.93} & \colorbox{gray!25}{82.49} \\
RhythmBERT (ours, single lead) & \colorbox{gray!25}{65.38} & \colorbox{gray!25}{70.71} & \colorbox{gray!25}{75.15} & \textbf{63.72} & \textbf{70.09} & \textbf{82.38} & \textbf{62.48} & \textbf{85.36} & \textbf{94.11} \\
\bottomrule
\end{tabular}
\end{table*}
\vspace{-2pt}
\subsection{Downstream Tasks Configuration}
\vspace{-2pt}
To rigorously assess the clinical relevance of our framework, we evaluate RhythmBERT across multiple downstream tasks (see Table \ref{tab:labels}) that include both highly prevalent arrhythmias and diagnostically challenging conditions. Common rhythms such as atrial fibrillation (AF, AFIB), conduction blocks (1AVB, LBBB, RBBB), and premature contractions (PAC, PVC) are well represented in the selected datasets and provide an essential benchmark for generalization to widespread heart conditions. At the same time, the inclusion of complex and often subtle abnormalities such as myocardial infarction (MI), ST-T deviations (STTC, STD, STE), supraventricular tachycardia (SVT), and hypertrophy (HYP) ensures that the model is tested against cases that are notoriously difficult even for expert interpretation. This design allows us to demonstrate not only broad applicability across clinically relevant scenarios but also the robustness of our approach in detecting nuanced pathologies that push the limits of automated ECG analysis.

\noindent\textbf{\underline{PTB-XL.}} A public dataset contains 21,799 12-lead ECGs from 18,869 patients \cite{wagner2020ptb}. Each ECG recording was sampled at 500 Hz and lasted for 10 seconds. Following the SCP-ECG proposal, diagnostic labels are categorized into 5 superclasses. We adhere to the official patient-stratified data splits for training, validation, and testing \cite{strodthoff2020deep}.

\noindent\textbf{\underline{CPSC2018.}} A public dataset contains 6,877 12-lead ECGs \cite{liu2018open}. Recordings were sampled at 500 Hz and vary in duration from 6 to 60 seconds. The dataset contains 9 distinct arrhythmia labels. We split the data into 70\%, 10\%, and 20\% for training, validation, and testing.

\noindent\textbf{\underline{Chapman-Shaoxing.}} A public dataset contains 10,646 12-lead ECG recordings \cite{zheng202012}. Each ECG recording was sampled at 500 Hz and lasted for 10 seconds. The dataset includes a total of 11 rhythm labels. We split the data into 70\%, 10\%, and 20\% for training, validation, and testing.

\noindent\textbf{\underline{Implementation.}} For downstream classification, we fine-tune the pre-trained model using Low-Rank Adaptation (LoRA) \cite{hu2022lora}. To explore the performance of our method under low-resource conditions, we conduct the fine-tuning using 1\%, 10\%, and 100\% of the training data for each task. The model is trained for up to 100 epochs with a batch size of 64. We use the AdamW optimizer with a weight decay of $1 \times 10^{-4}$. The learning rate schedule includes a 10\% warmup period and reaches a peak of $1 \times 10^{-4}$. Early stopping is employed. For all downstream tasks, we use the macro AUROC as the evaluation metric.  
\vspace{-4pt}
\subsection{Experimental Results}
\vspace{-1pt}
Table \ref{tab:results-auroc} presents the AUROC results of RhythmBERT compared to existing eSSL methods on three downstream datasets. On the CPSC2018 dataset, our model consistently outperforms both HeartLang and ST-MEM across all training data percentages. Specifically, it achieves an AUROC improvement of approximately 4.5\% over the second-best model when using 100\% of the training data. For the PTB-XL dataset, our method achieves competitive results, securing the second-best performance. We suspect this is because RhythmBERT is trained exclusively on Lead II, while the other two models use 12-lead data for training. Their models can capture more complex spatio-temporal information, giving them an advantage in this dataset. Notably, on the Chapman-Shaoxing dataset, our model shows superior performance. These results underscore RhythmBERT's strength in learning powerful representations for cardiac rhythm analysis, although trained exclusively on a single lead. 

Our results demonstrate that RhythmBERT achieves strong generalization not only across highly prevalent conditions such as atrial fibrillation, conduction blocks, and premature beats, but also in clinically challenging cases, including subtle ST-T abnormalities, myocardial infarction, and supraventricular tachycardia. This breadth of performance underscores the model’s capacity to capture physiologically meaningful representations that transfer robustly across both common and diagnostically difficult arrhythmias.\vspace{-4pt}

\subsection{Analysis of Latent Wave Representations}

\begin{figure}
  \centering
  \includegraphics[width=1\columnwidth]{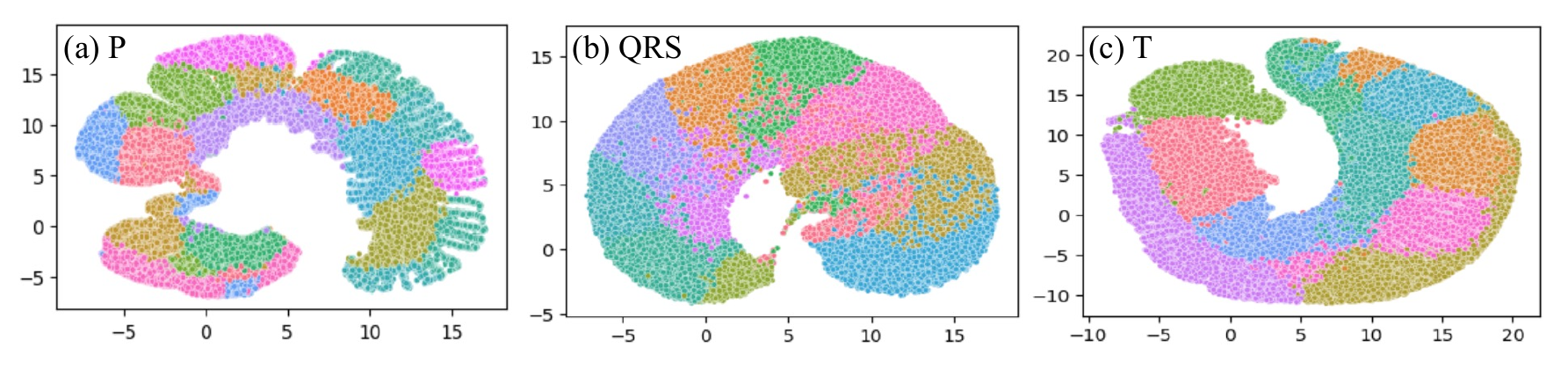}
  \vspace{-8pt}
  \caption{UMAP plots of k-means clusters for the latent representations of (a) P, (b) QRS, and (c) T waveforms, where the axes represent the two UMAP dimensions. Each point corresponds to one waveform segment, and colors indicate cluster assignments. For visualization clarity, a random subset of 50,000 samples is shown.}
  \label{fig:umap_kmeans}
\end{figure}

Fig.~\ref{fig:umap_kmeans} illustrates the UMAP plots of the P, QRS, and T clusters identified by k-means clustering. For the sake of computational efficiency, a random sample of 50,000 was chosen for this plot. The resulting projections highlight the structural patterns captured by k-means clustering.  \vspace{-2pt}

\section{Conclusion}
\label{sec:conclusion}
 \vspace{-1pt}
In this work, we introduce RhythmBERT, a novel self-supervised framework that advances the paradigm of treating waves as ``words'' and ECGs as ``sentences''. The discrete waveform tokens are fused with continuous morphological embeddings derived from a deep convolutional network, creating a unified representation of both rhythm and waveform morphology. Pre-training on approximately 800,000 unlabeled ECGs demonstrates the scalability of our approach, and extensive evaluations on downstream classification tasks confirm that RhythmBERT learns robust, transferable representations that are highly effective for detecting heart disease. Despite its focus on a single lead, our model demonstrates competitive and robust performance against other full-lead self-supervised learning models. Our results validate that this granular, physiologically-aligned approach offers a scalable pathway for representation learning in clinical time-series analysis. Future work will focus on extending this single-lead framework to multi-lead ECGs to model more complex spatio-temporal cardiac dynamics.

\vfill\pagebreak
\FloatBarrier      
\clearpage         
\balance
\bibliographystyle{IEEEbib}
\ninept
\bibliography{strings,refs}

\end{document}